\documentclass[10pt,twocolumn,letterpaper]{article}

\usepackage{wacv}              

\usepackage{graphicx}
\usepackage{amsmath}
\usepackage{amssymb}
\usepackage{booktabs}
\usepackage{multirow}

\usepackage[pagebackref,breaklinks,colorlinks]{hyperref}
\usepackage[accsupp]{axessibility}

\usepackage[capitalize]{cleveref}
\crefname{section}{Sec.}{Secs.}
\Crefname{section}{Section}{Sections}
\Crefname{table}{Table}{Tables}
\crefname{table}{Tab.}{Tabs.}

\def\wacvPaperID{1283} 
\def\confName{WACV}
\def\confYear{2024}

\begin{document}

\title{Painterly Image Harmonization via Adversarial Residual Learning}

\author{Xudong Wang, Li Niu\thanks{Corresponding author.}~,  Junyan Cao, Yan Hong, Liqing Zhang$^{*}$ \\
Department of Computer Science and Engineering, MoE Key Lab of Artificial Intelligence, \\
Shanghai Jiao Tong University\\
{\tt \small \{wangxudong1998,ustcnewly,Joy\_C1,hy2628982280,zhang-lq\}@sjtu.edu.cn}
}

\maketitle

\begin{abstract}
   Image compositing plays a vital role in photo editing. After inserting a foreground object into another background image, the composite image may look unnatural and inharmonious. When the foreground is photorealistic and the background is an artistic painting, painterly image harmonization aims to transfer the style of background painting to the foreground object, which is a challenging task due to the large domain gap between foreground and background. 
   In this work, we employ adversarial learning to bridge the domain gap between foreground feature map and background feature map. Specifically, we design a dual-encoder generator, in which the residual encoder produces the residual features added to the foreground feature map from main encoder. Then, a pixel-wise discriminator plays against the generator, encouraging the refined foreground feature map to be indistinguishable from background feature map.  
   Extensive experiments demonstrate that our method could achieve more harmonious and visually appealing results than previous methods.
\end{abstract}

\section{Introduction}
\label{sec:intro}

In many photo editing applications, it is often necessary to cut a foreground object from one image and overlay it on another background image, which is referred to as image composition \cite{niu2021making}.
However, when combining the foreground and background from different image sources to produce a composite image, the styles of foreground and background may be inconsistent, which would severely harm the quality of composite image.

\begin{figure}
  \centering
   \includegraphics[width=\linewidth]{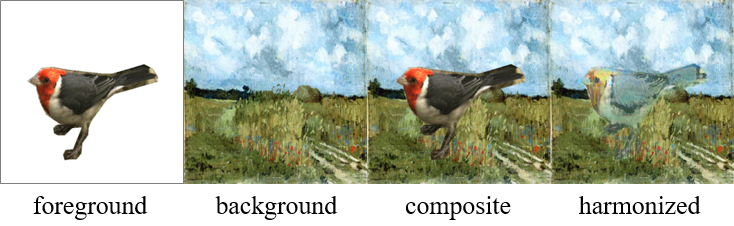}

   \caption{Example of painterly image harmonization. From left to right are foreground object, background image, composite image, and harmonized image.}
   \label{fig:example}
\end{figure}

When the foreground and background are both photographic images, the style mainly refers to illumination statistics, \emph{e.g.}, the foreground is captured in the daytime while the background is captured at night. To address the style inconsistency between foreground and background, image harmonization  \cite{tsai2017deep,cong2020dovenet,cun2020improving,ling2021region} aims to adjust the illumination statistics of foreground to be compatible with background, leading to a harmonious image. 
When the foreground is from a photographic image and the background is an artistic painting, the background style has the same meaning as in artistic style transfer \cite{gatys2016image,huang2017arbitrary,park2019arbitrary}, which includes color, texture, pattern, strokes, and so on. To address the style inconsistency between foreground and background, painterly image harmonization \cite{luan2018deep} aims to migrate the background style to the foreground, so that the stylized foreground is compatible with the background and naturally blended into the background. 

To the best of our knowledge, there are only few works on painterly image harmonization. To name a few, Luan et al. \cite{luan2018deep} proposed to update the composite foreground through iterative optimization process that minimizes the designed loss functions.
However, the method \cite{luan2018deep} relies on slow iterative optimization process, which imposes restrictions on real-time application.
Inspired by \cite{huang2017arbitrary}, Peng et al. \cite{peng2019element} introduced AdaIN \cite{huang2017arbitrary} to align the styles between foreground and background, which is trained with content loss and style loss.
The method \cite{peng2019element} runs much faster than \cite{luan2018deep}, but performs poorly when transferring the color and brush texture of artistic paintings.
Zhang et al. \cite{zhang2020deep} jointly optimized the proposed Poisson blending loss with the style and content loss, and reconstructed the blending region by iteratively updating the pixels. Analogous to \cite{luan2018deep}, the method \cite{zhang2020deep} is also very time-consuming. In summary, the existing painterly image harmonization methods are either time-consuming or weak in style transfer. Additionally, the image harmonization methods \cite{tsai2017deep,cong2020dovenet,cun2020improving,ling2021region} for photographic images are not suitable for our task (see Section \ref{sec:image_harmonization}) and the artistic style transfer methods \cite{gatys2016image,huang2017arbitrary,park2019arbitrary} have several limitations when applied to our task (see Section \ref{sec:artistic_style_transfer}).

One critical issue that hinders the performance of painterly image harmonization is the large domain gap between photographic foreground and painterly background. 
Considering that adversarial learning has been widely used to close the gap between different domains \cite{bousmalis2017unsupervised,tzeng2017adversarial}, we attempt to employ adversarial learning in the painterly image harmonization task. 
Actually, pixel-wise adversarial learning has been used in previous works \cite{huang2019temporally,kniaz2019point} from related fields (\emph{e.g.}, video harmonization, photo retouching). They use a discriminator to distinguish foreground pixels from background pixels in the output image, which can help strengthen the generator in an adversarial manner. 

\textbf{In this work, we apply similar idea to the feature maps in the generator, that is, employing adversarial learning to bridge the gap between foreground feature map and background feature map.}
Specifically, we propose a novel painterly image harmonization network that contains a dual-encoder generator (main encoder and residual encoder) and pixel-wise feature discriminators.
In the main encoder, we use pretrained VGG~\cite{VGG19} encoder to extract multiple layers of feature maps from composite image and background image. Then, we apply AdaIN \cite{huang2017arbitrary} to align the statistics between the foreground region in composite feature maps and the whole background feature maps, leading to stylized composite feature maps. 
To further reduce the domain gap between foreground and background, we also propose an extra residual encoder to learn residual features for each encoder layer. The learnt residual features are added to the foreground regions of stylized composite feature maps, leading to refined composite feature maps. 
Afterwards, for each encoder layer, our pixel-wise feature discriminator takes in the refined composite feature map and plays against our dual-encoder generator by telling disharmonious pixels from harmonious ones, which encourages the refined composite feature maps to be harmonious. Finally, the refined composite feature maps are delivered to the decoder to produce the harmonized image. We name our method as \textbf{P}ainterly \textbf{H}armonization via \textbf{A}dversarial \textbf{R}esidual \textbf{Net}work (PHARNet).

Following previous works \cite{luan2018deep,peng2019element}, we conduct experiments on COCO \cite{lin2014microsoft} and WikiArt \cite{nichol2016painter}, comparing with painterly image harmonization methods and artistic style transfer methods.
Our major contributions can be summarized as follows. 1) We are the first to introduce pixel-wise adversarial learning to harmonize feature maps. 2) We propose PHARNet equipped with novel dual-encoder generator and pixel-wise feature discriminator. 3) Extensive experiments on benchmark datasets prove the effectiveness of our network design. 

\section{Related Work}
\label{sec:releted}

\subsection{Image Harmonization} \label{sec:image_harmonization}

The goal of image harmonization is to harmonize a composite image by adjusting the illumination information of foreground to match that of background. 
Early traditional image harmonization methods \cite{song2020illumination,xue2012understanding,sunkavalli2010multi,lalonde2007using} tended to match low-level color or brightness information between foreground and background. 
After that, unsupervised deep learning methods \cite{zhu2015learning} were proposed to enhance the realism of harmonized image using adversarial learning. 
With the constructed large-scale dataset \cite{cong2020dovenet} containing paired training data, myriads of supervised deep learning approaches \cite{tsai2017deep,sofiiuk2021foreground,cong2022high,GiftNet,DucoNet} have been developed to advance the harmonization performance. To name a few, \cite{cun2020improving,hao2020image} designed various attention modules which are embedded in the network. \cite{cong2020dovenet,cong2021bargainnet} treated foreground and background as different domains, thus converting image harmonization task to domain translation task.  \cite{guo2021image,guo2021intrinsic} introduced intrinsic  decomposition to image harmonization task. More recently, \cite{cong2022high,ke2022harmonizer,liang2021spatial,xue2022dccf} integrated color transformation with deep learning network to achieve better performance. 
However, the well-behaved supervised image harmonization methods require pairs of training data, which are almost impossible to acquire in painterly image harmonization task.

\subsection{Painterly Image Harmonization}

When overlaying a photographic foreground onto a painterly background, the task is called painterly image harmonization. This task targeted at migrating the background style to the foreground and preserving the foreground content. As far as we are concerned, there only exist few works concentrating on painterly image harmonization task. The existing approaches \cite{luan2018deep,zhang2020deep,peng2019element} can be divided into optimization-based approaches \cite{luan2018deep,zhang2020deep} and feed-forward approaches \cite{peng2019element}.  The optimization-based approaches \cite{luan2018deep,zhang2020deep} iteratively optimize over the foreground region of input composite image to minimize the designed loss functions (\emph{e.g.}, content loss, style loss, Poisson loss), which is very inefficient. The feed-forward approaches \cite{peng2019element} pass the composite image through the network once and output the harmonized image, which is much more efficient than optimization-based methods.
PHDNet~\cite{cao2023painterly} performed image harmonization in both frequency domain and spatial domain. PHDiffusion~\cite{lu2023painterly} introduced diffusion model to painterly image harmonization. 

Our proposed method belongs to feed-forward approaches. Although adversarial learning has been used in \cite{peng2019element}, they perform image-level and region-level adversarial learning, which is quite different from our pixel-wise adversarial learning.
Moreover, \cite{peng2019element} tends to make the output images indistinguishable from artistic ones, but lacks the ability to match foreground style with background style.

\subsection{Artistic Style Transfer} \label{sec:artistic_style_transfer}
Artistic style transfer \cite{gatys2016image,li2017demystifying,huang2017arbitrary,liu2021adaattn,xu2021drb,zhang2022domain,chen2021dualast,chen2021artistic,wu2020efanet,lin2021drafting,svoboda2020two,jing2022learning,an2021artflow} renders a photo with a specific visual style by transferring style patterns from a given style image to a content image. Similar to painterly image harmonization, artistic style transfer methods can also be divided into optimization-based methods \cite{gatys2016image,li2017demystifying,kolkin2019style,du2020much} and feed-forward methods \cite{huang2017arbitrary,li2017universal,park2019arbitrary,liu2021adaattn,huo2021manifold,deng2022stytr2}. Artistic style transfer methods can be applied to painterly image harmonization task by transferring the style from background image to the whole content image and pasting the cropped stylized foreground on the background image. However, the foreground region is prone to be insufficiently stylized. Moreover, the pasted foreground may not be naturally blended into the background without considering the locality of compositing task.

\section{Our Method}
\label{sec:method}

\begin{figure*}
  \centering
   \includegraphics[width=0.95\linewidth]{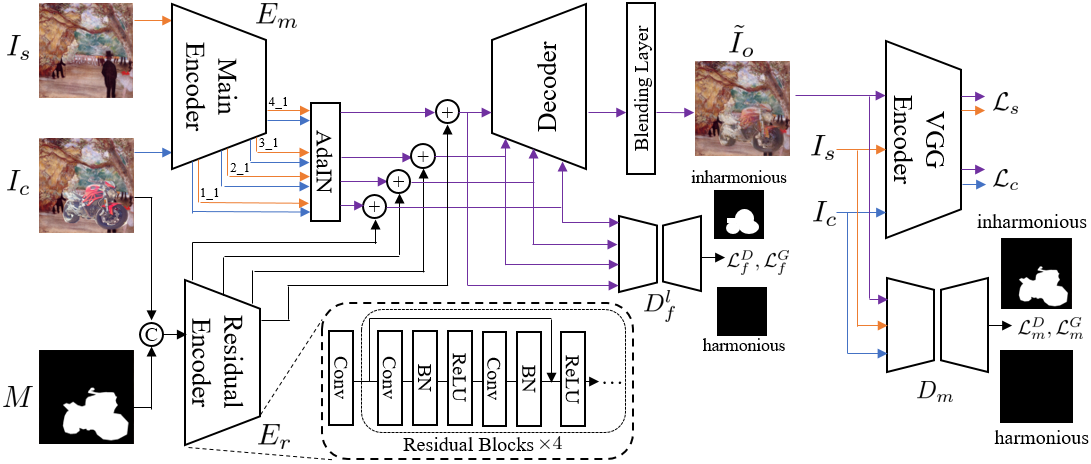}

   \caption{An overview of our painterly image harmonization network PHARNet. The network contains a dual-encoder generator $G$,  pixel-wise feature discriminators $D_f^l$, and a pixel-wise image discriminator $D_m$.}
   \label{fig:overview}
\end{figure*}

\subsection{Overview}
By pasting the foreground object from a photographic image on a painterly background image $I_s$, we can obtain the composite image $I_c$ with foreground mask $M$. 
The goal of painterly image harmonization is transferring the style of background image $I_s$ to the foreground object in the composite image $I_c$ while preserving the foreground content.

The overview of our network is shown in Figure \ref{fig:overview} , which contains a dual-encoder generator $G$,  pixel-wise feature discriminators $D_f^l$, and a pixel-wise image discriminator $D_m$. The dual-encoder generator $G$ consists of a main encoder $E_m$ and a residual encoder $E_r$. The generator $G$ takes in the background image $I_s$, the composite image $I_c$, and the foreground mask $M$, and generates a harmonized output image $\Tilde{I}_o$. In addition, we employ $L$ pixel-wise feature discriminators $\{D_f^l|_{l=1}^L\}$ and a pixel-wise image discriminator $D_m$ to play against $G$ by telling disharmonious pixels from harmonious ones. The pixel-wise feature discriminators $D_f^l$ are attached to multiple layers of feature maps in the generator, while the pixel-wise image discriminator $D_m$ is attached to the output image $\Tilde{I}_o$. Next, we will introduce each component in our network. 

\subsection{Dual-encoder Generator}
\label{sec:generator}
Our generator is composed of a main encoder $E_m$, a residual encoder $E_r$, and a decoder. The main encoder $E_m$ contains the first few layers (up to $ReLU 4\underline{\ }1$) of a pre-trained VGG-19 \cite{VGG19} and the decoder structure is symmetrical to the main encoder. We fix the main encoder $E_m$ when training our network. Following \cite{ronneberger2015u}, we add skip connections on $ReLU 1\underline{\ }1$, $ReLU 2\underline{\ }1$, and $ReLU 3\underline{\ }1$ to preserve the content details in the low-level feature maps.

At first, $E_m$ extracts $L=4$ layers of feature maps from the background image $I_s$ and the composite image $I_c$, leading to $\{F_s^l|_{l=1}^4\}$ and $\{F_c^l|_{l=1}^4\}$ from four encoder layers $ReLU 1\underline{\ }1$, $ReLU 2\underline{\ }1$, $ReLU 3\underline{\ }1$, and $ReLU 4\underline{\ }1$. For the $l$-th layer, we feed both feature maps $F_s^l$ and $F_c^l$ with the resized foreground mask $M^l$ to the AdaIN layer \cite{huang2017arbitrary} that aligns the statistics of the foreground region in $F_c^l$ with those of $F_s^l$, producing the stylized feature maps $F_a^l$:
\begin{eqnarray} \label{eqn:adain_layer}
    F_a^l=\!\!\!\!\!\!\!\!&&\Bigl(\sigma(F_s^l)\frac{F_c^l-\mu(F_c^l\circ M^l)}{\sigma(F_c^l\circ M^l)}+\mu(F_s^l)\Bigr)\circ M^l\\ \nonumber
    &&+F_c^l\circ(1-M^l),
\end{eqnarray}
where $\circ$ is Hadamard product, $\mu(\cdot)$ and $\sigma(\cdot)$ are the channel-wise mean and standard deviation of a certain region of a feature map.

Although the AdaIN operation in Eqn. \ref{eqn:adain_layer} roughly aligns the composite foreground with the background image, the domain gap between foreground and background in $F_a^l$ may still exist. Therefore, we attempt to refine the foreground details in the stylized feature maps to further reduce the domain gap. To this end, we design a residual encoder $E_r$ to learn multiple layers of residual features that are added to the foreground regions of stylized feature maps $\{F_a^l|_{l=1}^4\}$.

Our residual encoder $E_r$ takes the concatenation of the composite image $I_c$ and the foreground mask $M$ as input. 
We employ four residual blocks to learn four layers of residual features. 
All residual blocks share the identical structure, that is, two convolutional filters followed by batch-normalization layer and ReLU activation. For the $l$-th layer, the learned residual features $F_r^l$, \emph{i.e.}, the output from the $l$-th residual block, are added to the foreground region in the stylized feature map $F_a^l$, leading to refined feature map $\Tilde{F}_a^l$:
\begin{equation}
    \Tilde{F}_a^l=F_a^l+F_r^l\circ M^l.
\end{equation}

Then, multiple layers of refined feature maps are delivered to the decoder through bottleneck or skip connection to generate the output image $I_o$.
Afterwards, inspired by \cite{sofiiuk2021foreground}, we adopt a blending layer to blend $I_o$ with the background image $I_s$.
In particular, we feed the concatenation of the final decoder feature map and the foreground mask $M$ to the blending layer \cite{sofiiuk2021foreground}, generating a soft mask $\Tilde{M}$. At last, we blend the output image $I_o$ with the background image $I_s$ using $\Tilde{M}$ to obtain the final harmonized image $\Tilde{I}_o$:
\begin{equation}
    \Tilde{I}_o=I_o\circ\Tilde{M}+I_s\circ(1-\Tilde{M}).
\end{equation}

\subsection{Pixel-wise Feature Discriminator}
To supervise the learned residual features and mitigate the foreground-background domain gap in the refined feature maps, 
we employ pixel-wise adversarial learning to encourage the foreground pixels to be indistinguishable from the background pixels in the refined feature maps. 

We attach a pixel-wise feature discriminator $D_f^l$ to the $l$-th layer of refined feature map $\Tilde{F}_a^l$. $D_f^l$ aims to distinguish inharmonious pixels from harmonious pixels and assign a class label to each pixel in the feature map. Considering the output format, we adopt encoder-decoder architecture for $D_f^l$, which produces a mask. We use $D_f^l(\Tilde{F_a^l})$ to denote the discriminator output for $\Tilde{F_a^l}$. $D_f^l(\Tilde{F_a^l})$ should be close to $M^l$ so that the discriminator is guided to distinguish the foreground pixels from background pixels, in which the foreground pixels are labeled as $1$ and the background pixels are labeled as $0$. We also feed the feature map $F_s^l$ of background image into the discriminator. Since there are no inharmonious pixels in the background image, all pixels should be labeled as $0$. Therefore, the loss function to train the discriminator $D_f^l$ can be written as
\begin{equation} \label{eqn:loss_ldf}
        L^D_f=\sum_{l=1}^4\Vert D_f^l(\Tilde{F}_a^l)-M^l\Vert_2^2 + \sum_{l=1}^4\Vert D_f^l(F_s^l)\Vert_2^2.
\end{equation}

When training the generator $G$, we expect that the foreground pixels are indistinguishable from the background pixels in the refined feature maps, that is, all pixels should be labeled as $0$. Thus, the loss function for $D_f^l$ can be written as
\begin{equation} \label{eqn:loss_lgf}
    L^G_f=\sum_{l=1}^4\Vert D_f^l(\Tilde{F}_a^l)\Vert_2^2.
\end{equation}

Note that, unlike the commonly used global discriminator which classifies an image or a feature map to be real or fake as a whole, our pixel-wise feature discriminator learns to classify each pixel-wise feature vector separately. 

\subsection{Other Losses}
In this section, we introduce the remaining losses imposed on the final harmonized image $\Tilde{I}_o$.

We employ the style loss \cite{huang2017arbitrary} to ensure that the style of foreground object is close to that of background image:
\begin{equation}
    \begin{split}
        L_s=\sum_{l=1}^4 \Vert\mu(\Psi^l(\Tilde{I}_o)\circ M^l)-\mu(\Psi^l(I_s))\Vert_2^2 + \\
        \sum_{l=1}^4 \Vert\sigma(\Psi^l(\Tilde{I}_o)\circ M^l)-\sigma(\Psi^l(I_s))\Vert_2^2,
    \end{split}
\end{equation}
in which $\Psi^{l}$ denotes the $l$-th \emph{ReLU-l\_1} layer in the pretrained VGG-19 encoder.

We also employ the content loss \cite{gatys2016image} to enforce the harmonized image to retain the content of the foreground object:
\begin{equation}
    L_c=\Vert\Psi^4(\Tilde{I}_o)-\Psi^4(I_c)\Vert_2^2.
\end{equation}

Inspired by \cite{huang2019temporally}, we also apply pixel-wise adversarial learning to the harmonized image $\Tilde{I}_o$. Specifically, we train a pixel-wise image discriminator $D_m$ to distinguish inharmonious pixels from harmonious pixels by minimizing the loss $L^D_m$, while the generator strives to make the foreground pixels indistinguishable from background pixels by minimizing the loss $L^G_m$. The definitions of $L^D_m$  and $L^G_m$ are similar to $L^D_f$ in Eqn. \ref{eqn:loss_ldf} and $L^G_f$ in Eqn. \ref{eqn:loss_lgf} except the input, so we omit the details here.

In summary, the total loss function for training the generator $G$ is
\begin{equation}
    L_G= L_c+ L_s+ L^G_f+L^G_m.
\end{equation}
The total loss function for training the discriminators $\{D^l_f|_{l=1}^4\}$ and $D_m$ is
\begin{equation}
    L_D= L^D_f+ L^D_m.
\end{equation}
Under the adversarial learning framework, we update the generator and the discriminators alternatingly.

\section{Experiments}
\label{sec:experiments}

\subsection{Datasets}
Following previous works on painterly image harmonization \cite{luan2018deep,peng2019element,zhang2020deep}, we conduct experiments on COCO \cite{lin2014microsoft} and WikiArt \cite{nichol2016painter} datasets. 
COCO is a large-scale dataset of 123,287 images, which have instance segmentation annotations for the objects from 80 categories.
Wikiart is a large-scale digital art dataset which contains 81444 images from 27 different styles.
In this work, we use the images from WikiaArt dataset as painterly background images and extract photographic foreground objects from COCO dataset using the provided instance segmentation masks. 
We randomly choose a segmented object whose area ratio in the original image is in the range of [0.05, 0.3], and paste it onto a randomly selected painting background, producing an inharmonious composite image. 
We follow the training and test split of COCO and WikiArt as \cite{tan2017artgan}, based on which we obtain 57,025 (\emph{resp.}, 24,421) background images and 82783 (\emph{resp.}, 40504) foreground objects for training (\emph{resp.}, testing).
\subsection{Implementation Details}

The overall architecture of our network has been described in Section \ref{sec:method}. 
For the residual encoder $E_r$, we use four residual blocks \cite{he2016deep} to learn the residual features.
All residual blocks share the identical structure, that is, two convolutional filters followed by batch-normalization layer and ReLU activation. 
The pixel-wise feature discriminators $D_f^l$ are small-scale auto-encoders consisting of downsample (DS) and upsample (US) blocks.
For $l\in\{1,2\}$, $D_f^l$ contains three DS blocks and three US blocks.
For $l\in\{3,4\}$, $D_f^l$ contains two DS blocks and two US blocks.
Each DS block contains a convolutional layer with kernel size being $4$ and stride being $2$, a batch normalization layer, and a LeakyReLU activation sequentially.
Each US block contains an upsampling layer with scale factor being $2$, a reflection padding layer, a convolutional layer with kernel size being $3$ and stride being $1$, a batch normalization layer, and a ReLU layer.
The pixel-wise image discriminator $D_m$ is also built upon DS blocks and US blocks as used in $D_f^l$. 
For $D_m$, we employ seven DS blocks and seven US blocks.

Our network is implemented with Pytorch 1.10.2 and trained using Adam optimizer with learning rate of $2e-4$ on ubuntu 18.04 LTS operating system, which has 32GB of
memory, Intel Core i7-9700K CPU, and two GeForce GTX 2080 Ti GPUs. We resize the input images to $256\times256$ in the training stage. However, our network can be applied to the test images of arbitrary size due to the fully convolutional network structure.
\subsection{Baselines}
There are two groups of methods which can be applied to our task: painterly image harmonization \cite{luan2018deep,peng2019element,zhang2020deep} and artistic style transfer \cite{huang2017arbitrary,liu2021adaattn}.

For the first group of methods, we compare with Deep Image Blending \cite{zhang2020deep} (“DIB” for short), Deep Painterly Harmonization \cite{luan2018deep} (“DPH” for short), and E2STN \cite{peng2019element}. 
We also include traditional image blending method Poisson Image Editing \cite{perez2003poisson} (“Poisson” for short) for comparison.

For the second group of methods, they were originally proposed to migrate the style of an artistic image to a complete photographic image, so some modifications are required to adapt them to our task. 
In particular, we first migrate the style of background image to the photographic image containing the foreground object, using the artistic style transfer methods.  
Then, we segment the stylized foreground object and overlay it on the background image to obtain a harmonized image. Since there are myriads of artistic style transfer methods, we choose several iconic or recent works for comparison: WCT \cite{li2017universal}, AdaIN \cite{huang2017arbitrary}, SANet \cite{park2019arbitrary}, AdaAttN \cite{liu2021adaattn}, and StyTr2 \cite{deng2022stytr2}.

\subsection{Qualitative Analysis}
\begin{figure*}
  \centering
   \includegraphics[width=0.97\linewidth]{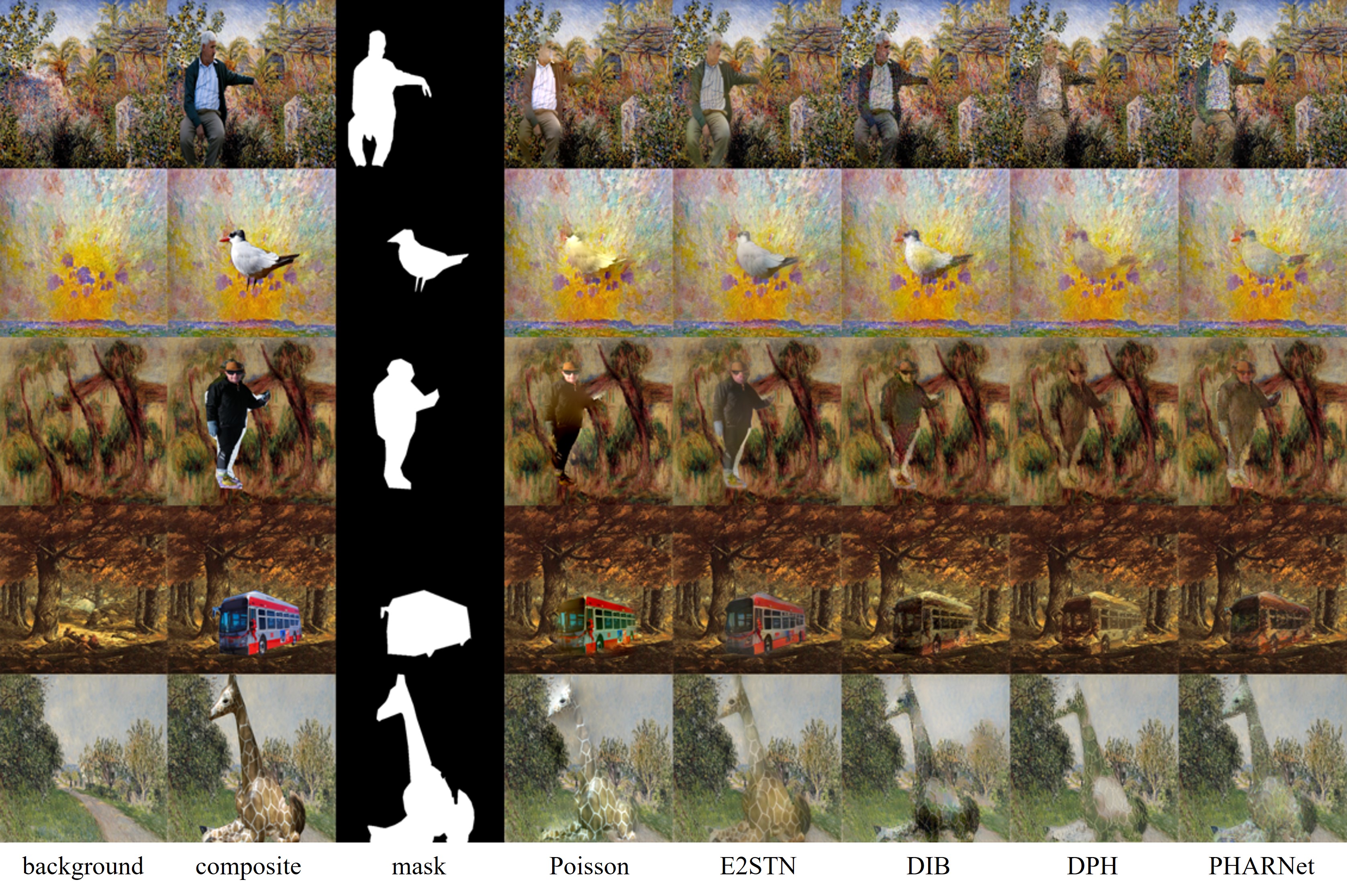}

   \caption{From left to right, we show the background image, composite image, foreground mask, the harmonized results of Poisson \cite{perez2003poisson}, E2STN \cite{peng2019element}, DIB \cite{zhang2020deep}, DPH \cite{luan2018deep}, and our PHARNet.}
   \label{fig:results_painterly_harmonization}
\end{figure*}

We show the comparison with the first group of baselines in Figure \ref{fig:results_painterly_harmonization} and the comparison with the second group of baselines in Figure \ref{fig:results_style_transfer}. More visualization results could be found in Supplementary. 

As shown in Figure \ref{fig:results_painterly_harmonization}, Poisson \cite{perez2003poisson} can smoothen the boundary between foreground and background, but the foreground content is severely distorted (\emph{e.g.}, row 2, 5). 
DIB \cite{zhang2020deep} and E2STN \cite{peng2019element} preserve the foreground content well, but the foreground style is not very close to background style (\emph{e.g.}, E2STN in row 4, DIB in row 2) and the harmonized foreground may be corrupted (\emph{e.g.}, DIB in row 5). DPH  \cite{luan2018deep} is a competitive baseline, which can achieve good harmonized results in some cases. However, the content structure and foreground boundary might be damaged or blurred (\emph{e.g.}, row 2, 5). In comparison, our method can preserve the content structure, sharp boundaries, and rich details  (\emph{e.g.}, human face/clothes in row 1, 3 and the patterns on the giraffe body in row 5). In the meanwhile, the foreground is sufficiently stylized and harmonious with the background. Interestingly, without suppressing the stylization effect, our method can also maintain the color distribution of foreground (\emph{e.g.}, white-and-red car in row 4), while other methods either understylize the foreground or lose partial color distribution information. 

\begin{figure*}
  \centering
   \includegraphics[width=0.97\linewidth]{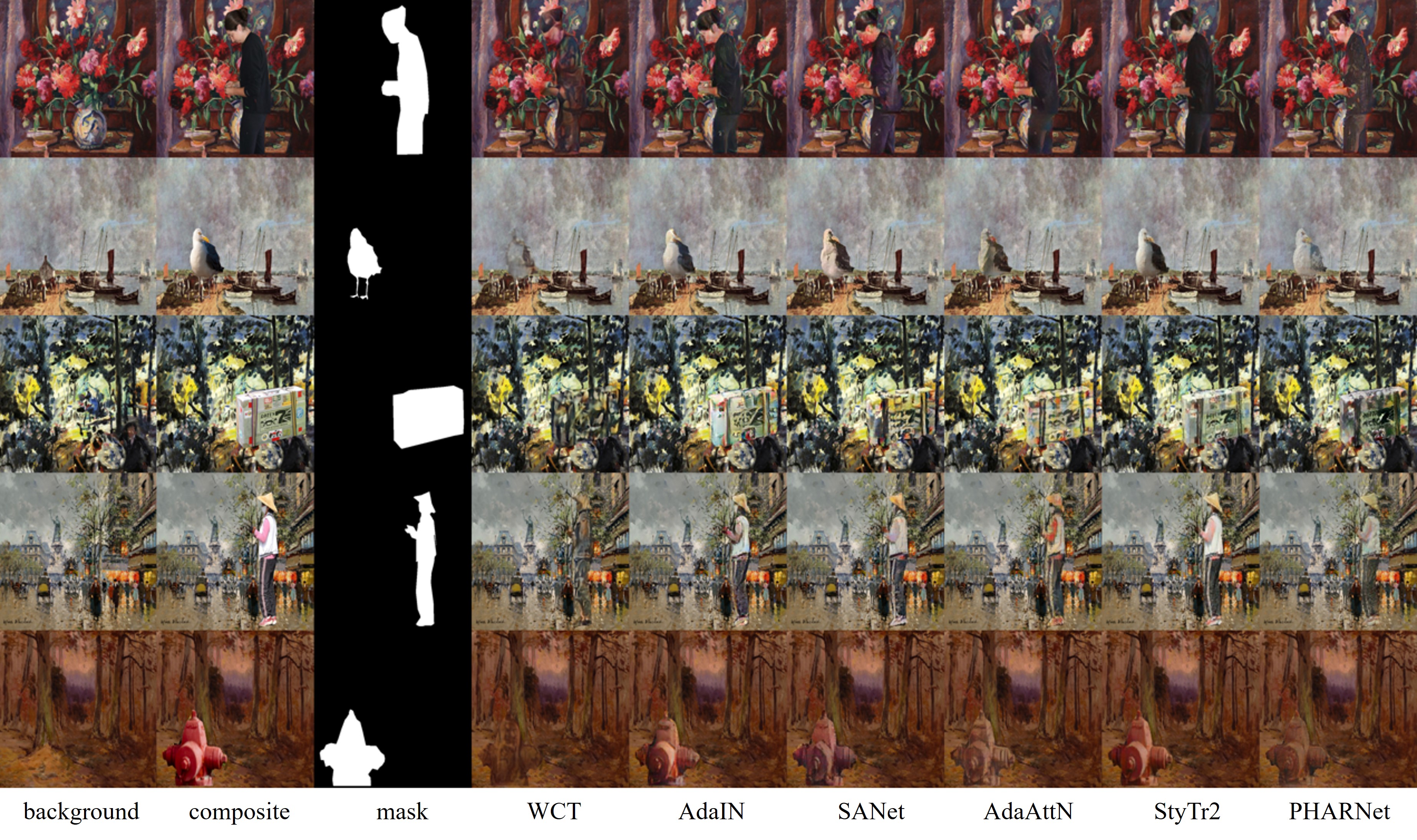}

   \caption{From left to right, we show the background image, composite image, foreground mask, the harmonized results of WCT \cite{li2017universal}, AdaIN \cite{huang2017arbitrary}, SANet \cite{park2019arbitrary}, AdaAttN \cite{liu2021adaattn}, StyTr2 \cite{deng2022stytr2}, and our PHARNet.}
   \label{fig:results_style_transfer}
\end{figure*}

\begin{table*}
  \centering
  \begin{tabular}{c|c|c|c|c|c|c}
    \toprule
    Method & SANet \cite{park2019arbitrary}  &  AdaAttN \cite{liu2021adaattn} &  StyTr2 \cite{deng2022stytr2} & E2STN \cite{peng2019element} &  DPH \cite{luan2018deep} & 
    PHARNet \\
    \hline
    BT-Score & -1.8757 & -1.0406 & -0.3891  & 0.4677 & 0.7814 & 2.0562 \\
    \hline
    Time(s) & 0.0097 & 0.0115 & 0.0504 & 0.0078 & 270.96 &  0.0223 \\
    \bottomrule
  \end{tabular}
  \caption{The BT-score and inference time of different methods.}
  \label{tab:quantitative_results}
\end{table*}

As shown in Figure \ref{fig:results_style_transfer}, since the style transfer methods do not focus on stylizing the foreground region, the foreground may not be adequately stylized (\emph{e.g.}, AdaIN and StyTr2 in row 2) and the content structure of foreground may be destroyed (\emph{e.g.}, WCT in row 4, 5). Besides, since style transfer methods do not consider the location of foreground in the composite image, the stylized foreground may be incompatible with the surrounding background. In contrast, our method is able to transfer the style and retain the content structure, leading to more visually appealing results. The stylized foregrounds are harmonious with backgrounds, as if they originally exist in the paintings.

\subsection{User Study}
We randomly select $100$ foreground objects and $100$ background images to generate $100$ composite images for user study. 
We compare with $5$ representative baselines SANet \cite{park2019arbitrary}, AdaAttN \cite{liu2021adaattn}, StyTr2 \cite{deng2022stytr2}, E2STN \cite{peng2019element}, DPH \cite{luan2018deep}. Specifically, for each composite image, we can obtain $6$ harmonized images produced by $6$ methods, based on which $2$ images are selected to construct an image pair. 
Provided with $100$ composite images, we can construct in total $1500$ image pairs.
Then, we ask $50$ annotators to observe one image pair at a time and pick the better one. 
At last, we gather $30,000$ pairwise results and  calculate the overall ranking of all methods using Bradley-Terry (B-T) model \cite{bradley1952rank,lai2016comparative}. 
As shown in Table \ref{tab:quantitative_results}, our method achieves the highest B-T score.

\subsection{Efficiency Comparison}
We compare the inference time between our method and baseline methods in Table~\ref{tab:quantitative_results}.
We test the inference speed of all methods on one GeForce GTX 2080 Ti GPU, with input image size $256 \times 256$, and average the results over $100$ test images. We observe that DPH is the slowest method because DPH is an optimization-based method which requires iterative optimization process. StyTr2~\cite{deng2022stytr2} is also very slow due to the Transformer network structure. Our method is relatively efficient and the inference speed is acceptable for real-time applications.

\begin{table}
  \centering
  \begin{tabular}{c|c|c|c|c}
    \toprule
    Method  & $E_r$ & $D_f^l$ & $D_m$ & B-T score \\
    \midrule
    V1  &  &  & & -4.4103  \\
    V2  &  &  & \checkmark & -0.6537  \\
    V3  & \checkmark &  & \checkmark & 1.4343 \\
    V4  & \checkmark & \checkmark & \checkmark & 3.6297 \\
    \bottomrule
  \end{tabular}
  \caption{The B-T score of our different network structure. $E_r$ refers to the residual encoder. $D_f^l$ refers to the pixel-wise feature discriminators. $D_m$ refers to the pixel-wise image discriminator.}
  \label{tab:ablation_studies}
\end{table}

\subsection{Ablation Studies}

In this section, we investigate the effectiveness of each component in our method. We first remove all discriminators and the residual encoder, and obtain a basic network with multi-scale AdaIN, which is referred to as V1 in Table \ref{tab:ablation_studies}. Then, we add pixel-wise image discriminator $D_m$, which is referred to as V2. Furthermore, we add the residual encoder $E_r$ to form the dual-encoder generator, which is referred to as V3. Finally, we apply pixel-wise feature discriminators and reach our full-fledged method, which is referred to as V4. 

\begin{figure*}
  \centering
   \includegraphics[width=0.83\linewidth]{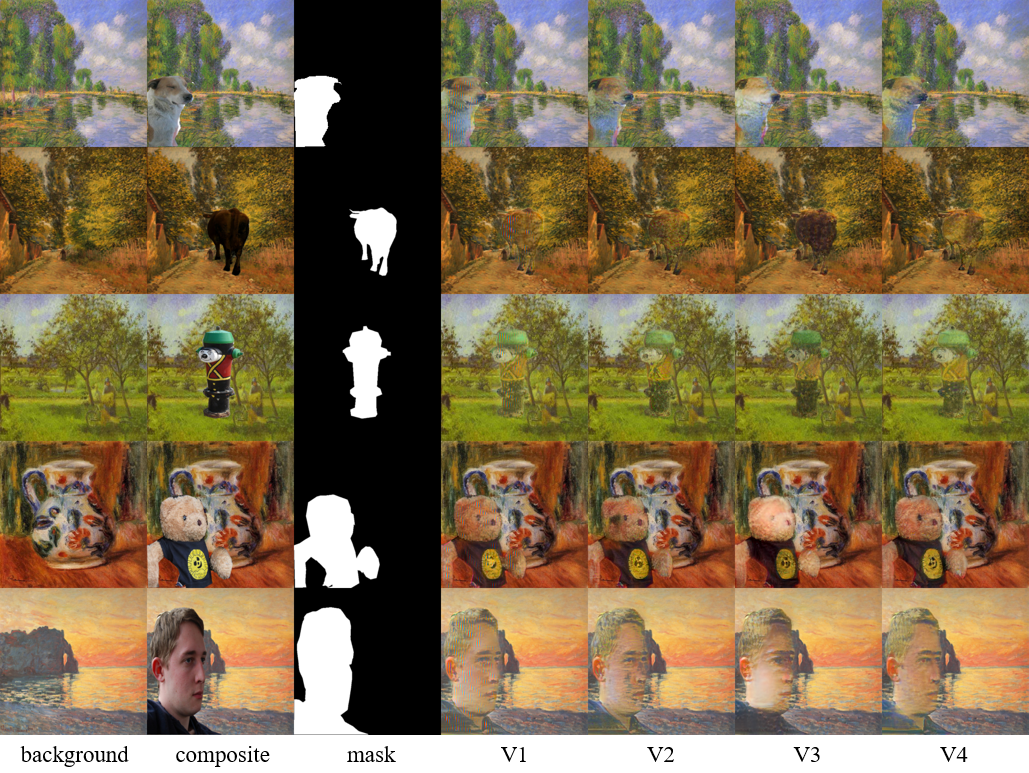}

   \caption{From left to right, we show the background image, composite image, foreground mask, the harmonized results of V1, V2, V3, V4 (full method).}
   \label{fig:results_ablation_studies}
\end{figure*}

We show the harmonized results of ablated versions in Figure \ref{fig:results_ablation_studies}. It can be seen that the harmonized results of V1 have many strip artifacts, which significantly harms the quality of harmonized results. After using the pixel-wise image discriminator $D_m$ in V2, the strip artifacts can be removed. Nevertheless, the harmonized results of V2 may still have some other types of artifacts (\emph{e.g.}, row 4, row 5) and unsatisfactory details. After adding the residual features without the guidance of pixel-wise feature discriminator, the harmonized foregrounds of V3 may have distorted content (\emph{e.g.}, row 5) and look incompatible with the background (\emph{e.g.}, row 4). After applying pixel-wise feature discriminators to the refined feature map in V4, the learnt residual features become more reasonable and the harmonized results become more visually appealing. Compared with the ablated versions, the results of V4 have fully-transferred style, well-preserved content structure, and meaningful details (\emph{e.g.}, dog eye in row 1, suspender in row 3). 

The clear advantage of V4 could be attributed to the residual features and pixel-wise adversarial learning. The residual features, which are added to the foreground region of stylized feature map, could repair the content structure and enhance the style representations, leading to the refined feature map with improved quality. Moreover,  the pixel-wise feature discriminator plays against the dual-encoder generator by telling disharmonious pixels from harmonious ones. Such pixel-wise adversarial learning encourages the refined foreground feature map to be indistinguishable from background feature map, so that the foreground is more harmonious with the surrounding background. 

Similar to Table \ref{tab:quantitative_results}, we also conduct user study to compare different ablated versions. The results are summarized in Table \ref{tab:ablation_studies}, which again demonstrate the superiority of our full method.

\section{Discussion on Limitation}
\label{sec:failure}
\begin{figure}[t]
    \centering
    \begin{subfigure}{0.95\linewidth}
        \includegraphics[width=\linewidth]{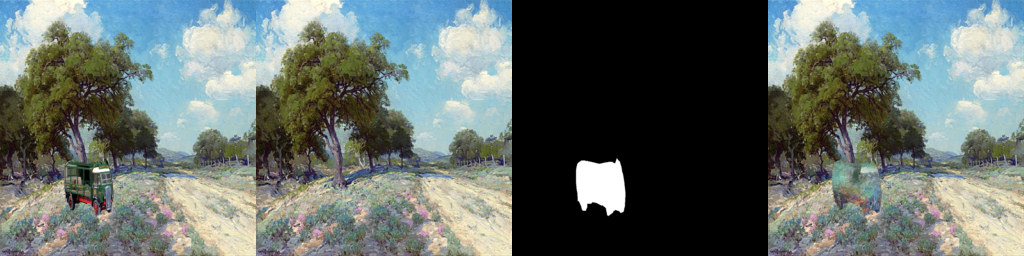}
    \end{subfigure}
    \begin{subfigure}{0.95\linewidth}
        \includegraphics[width=\linewidth]{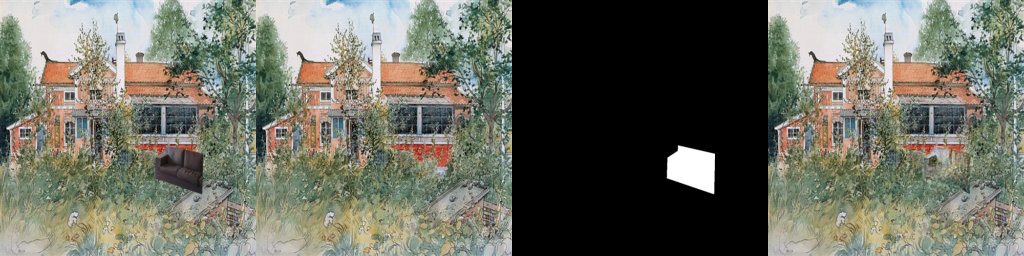}
    \end{subfigure}
    \begin{subfigure}{0.95\linewidth}
        \captionsetup{labelformat=empty}
        \begin{subfigure}{0.24\linewidth}
            \caption{composite}
        \end{subfigure}
        \begin{subfigure}{0.24\linewidth}
            \caption{background}
        \end{subfigure}
        \begin{subfigure}{0.24\linewidth}
            \caption{mask}
        \end{subfigure}
        \begin{subfigure}{0.24\linewidth}
            \caption{ours}
        \end{subfigure}
    \end{subfigure}
    \caption{Example failure cases of our method.}
    \label{fig:failure}
\end{figure}

Although our method can generally produce visually appealing results, there still exist some challenging cases in which our method may fail to produce satisfactory results. For example, as shown in Figure \ref{fig:failure}, when the foreground objects are very small, our method may fail in retaining the foreground content information and produce poor harmonized results.

\section*{Acknowledgement}
The work was supported by the National Natural Science Foundation of China (Grant No. 62076162), the Shanghai Municipal Science and Technology Major/Key Project, China (Grant No. 2021SHZDZX0102, Grant No. 20511100300). 

{\small
\bibliographystyle{ieee_fullname}
\bibliography{main.bbl}
}

\end{document}


\title{Supplementary for Painterly Image Harmonization via Adversarial Residual Learning}

\author{Xudong Wang, Li Niu\thanks{Corresponding author.}~,  Junyan Cao, Yan Hong, Liqing Zhang$^{*}$ \\
Department of Computer Science and Engineering, MoE Key Lab of Artificial Intelligence, \\
Shanghai Jiao Tong University\\
{\tt \small \{wangxudong1998,ustcnewly,Joy\_C1,hy2628982280,zhang-lq\}@sjtu.edu.cn}
}

\maketitle

In this supplementary, we will first provide more visualization results of different methods in Section \ref{sec:baseline}.
Then, we will analyze the impact of adding residual features to different layers in Section \ref{sec:residual}.
Then, we will show the harmonized results of the same foreground pasted on different backgrounds in Section \ref{sec:foreground}.
We will compare different adversarial losses in Section \ref{sec:adversarial} and show the results of multiple foregrounds in Section \ref{sec:multifore}.

\section{Visual Comparison with Baselines}
\label{sec:baseline}
\begin{figure*}[t]
  \centering
    \begin{subfigure}{\linewidth}
        \includegraphics[width=\linewidth]{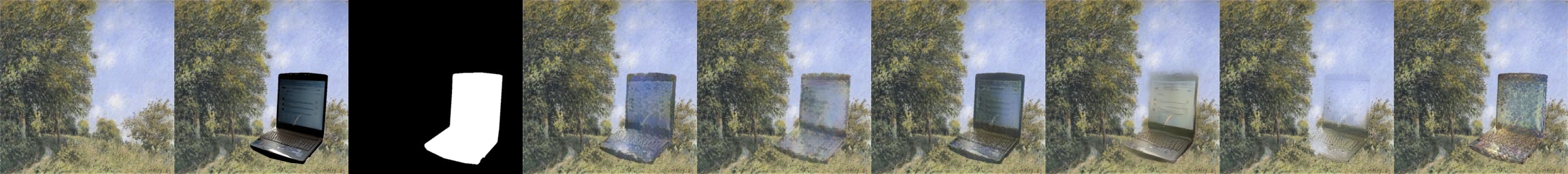}
    \end{subfigure}
    \begin{subfigure}{\linewidth}
        \includegraphics[width=\linewidth]{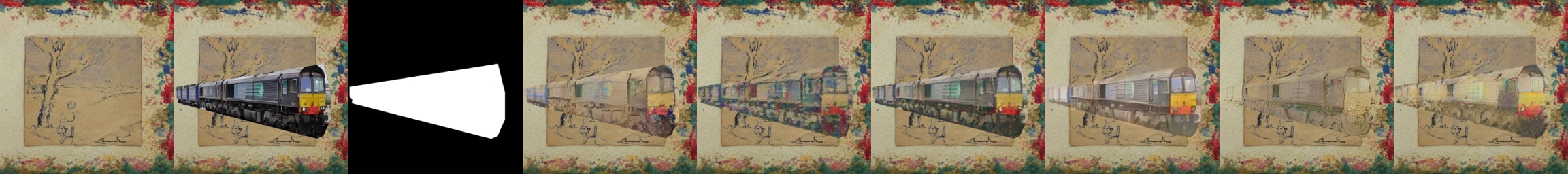}
    \end{subfigure}
    \begin{subfigure}{\linewidth}
        \includegraphics[width=\linewidth]{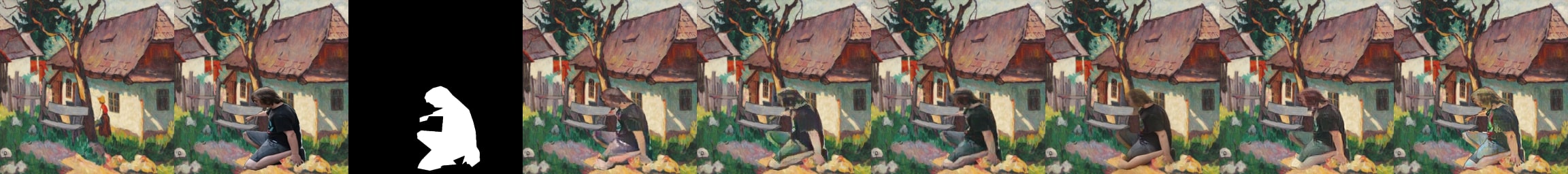}
    \end{subfigure}
    \begin{subfigure}{\linewidth}
        \includegraphics[width=\linewidth]{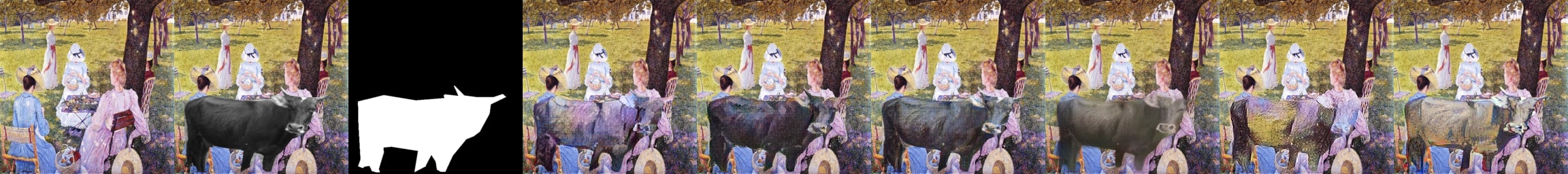}
    \end{subfigure}
    \begin{subfigure}{\linewidth}
        \includegraphics[width=\linewidth]{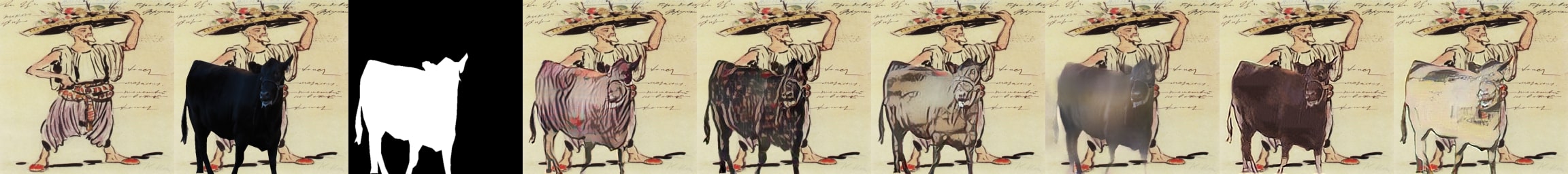}
    \end{subfigure}
    \begin{subfigure}{\linewidth}
        \includegraphics[width=\linewidth]{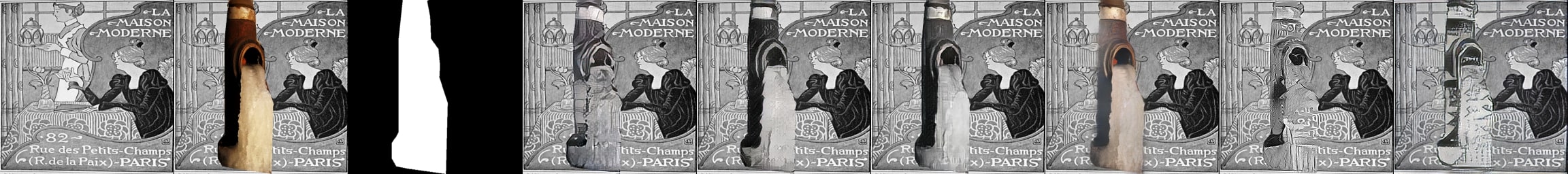}
    \end{subfigure}
    \begin{subfigure}{\linewidth}
        \includegraphics[width=\linewidth]{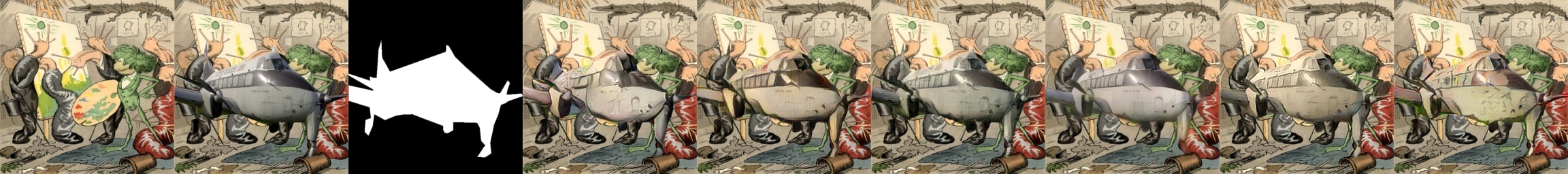}
    \end{subfigure}
    \begin{subfigure}{\linewidth}
        \includegraphics[width=\linewidth]{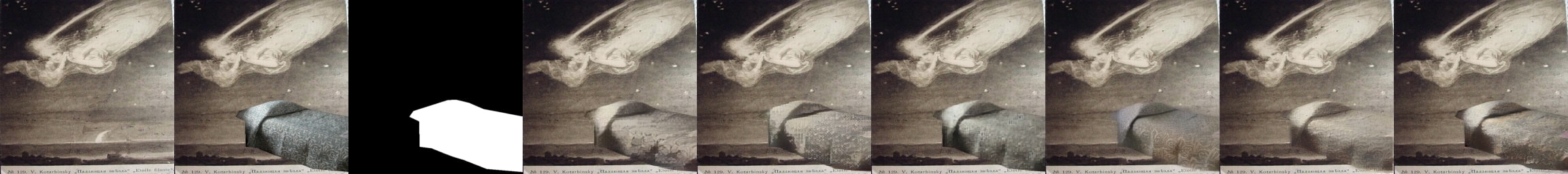}
    \end{subfigure}
    \begin{subfigure}{\linewidth}
        \includegraphics[width=\linewidth]{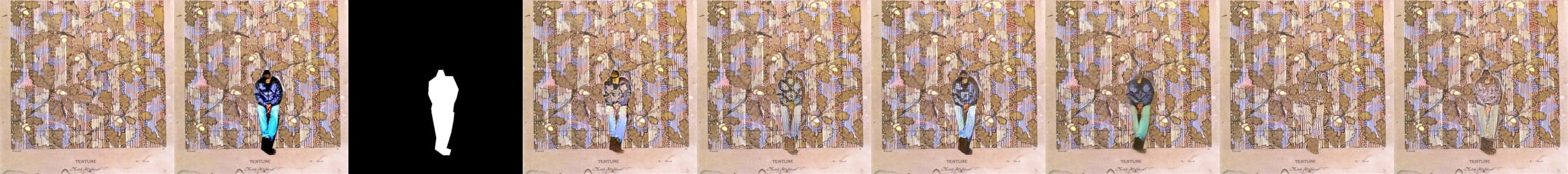}
    \end{subfigure}
    \begin{subfigure}{\linewidth}
        \includegraphics[width=\linewidth]{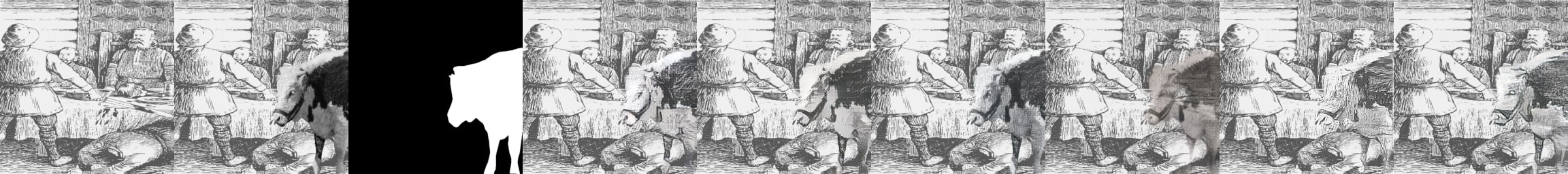}
    \end{subfigure}
    \begin{subfigure}{\linewidth}
        \captionsetup{labelformat=empty}
        \begin{subfigure}{0.106\linewidth}
            \caption{background}
        \end{subfigure}
        \begin{subfigure}{0.106\linewidth}
            \caption{composite}
        \end{subfigure}
        \begin{subfigure}{0.106\linewidth}
            \caption{mask}
        \end{subfigure}
        \begin{subfigure}{0.106\linewidth}
            \caption{SANet}
        \end{subfigure}
        \begin{subfigure}{0.106\linewidth}
            \caption{AdaAttN}
        \end{subfigure}
        \begin{subfigure}{0.106\linewidth}
            \caption{StyTr2}
        \end{subfigure}
        \begin{subfigure}{0.106\linewidth}
            \caption{E2STN}
        \end{subfigure}
        \begin{subfigure}{0.106\linewidth}
            \caption{DPH}
        \end{subfigure}
        \begin{subfigure}{0.106\linewidth}
            \caption{PHARNet}
        \end{subfigure}
    \end{subfigure}
  \caption{From left to right, we show the background image, composite image, foreground mask, the harmonized results of SANet \cite{park2019arbitrary}, AdaAttN \cite{liu2021adaattn}, StyTr2 \cite{deng2022stytr2}, E2STN \cite{peng2019element}, DPH \cite{luan2018deep}, and our PHARNet.}
  \label{fig:baseline}
\end{figure*}

We choose the competitive baselines SANet \cite{park2019arbitrary}, AdaAttN \cite{liu2021adaattn}, StyTr2 \cite{deng2022stytr2}, E2STN \cite{peng2019element}, DPH \cite{luan2018deep} from two groups of baselines, in which E2STN and DPH are from painterly image harmonization group while the rest are from the artistic style transfer group.
In Figure \ref{fig:baseline}, we show the harmonized results generated by baseline methods and our method.
Compared with these baselines, our method can successfully preserve the foreground content and transfer style from background image.

For example, our method can preserve fine-grained details (\emph{e.g.}, row 1, 2) and sharp contours (\emph{e.g.}, row 8) while transferring the style, which achieves a better balance between style and content. In contrast, the baseline methods may under-stylize the foreground so that the foreground is not harmonious with background, or severely distort the content structure so that the foreground is hardly recognizable.  In some challenging cases, our method can better transfer the style (\emph{e.g.}, color, texture) and obtain more visually appealing results (\emph{e.g.}, row 4, 5, 6, 9), while the baselines fail to make the foreground style compatible with the background.
Overall, in our harmonized images, the foreground is properly stylized and harmonious with the background so that the whole image appears to be an intact artistic painting. 

\section{Adding Residuals to Different Layers}
\label{sec:residual}
As described in Section 3.2 in the main paper, we employ four residual blocks to learn four layers of residual features.
For the $l$-th layer of the main encoder, the learned residual feature $F_r^l$, which is the output from the $l$-th residual block, is added to the foreground region in the stylized feature map $F_a^l$, leading to the refined feature map $\Tilde{F_a^l}$.

\begin{figure}[t]
    \centering
    \begin{subfigure}{0.95\linewidth}
        \includegraphics[width=\linewidth]{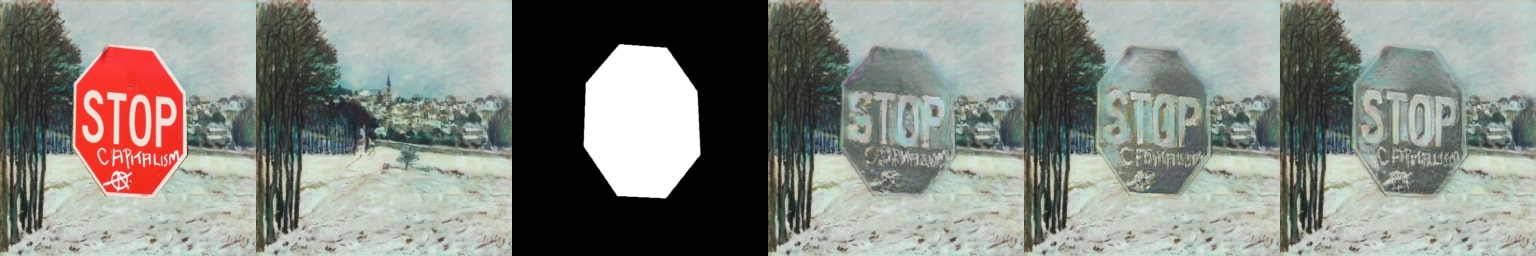}
    \end{subfigure}
    \begin{subfigure}{0.95\linewidth}
        \includegraphics[width=\linewidth]{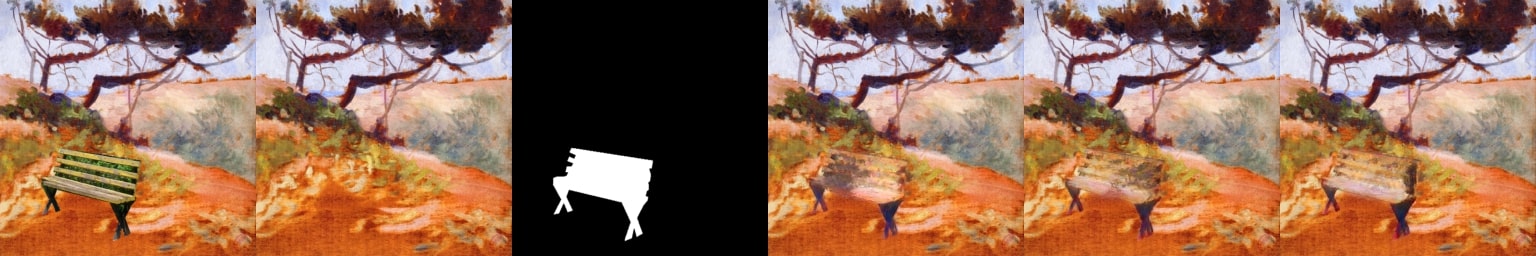}
    \end{subfigure}
    \begin{subfigure}{0.95\linewidth}
        \includegraphics[width=\linewidth]{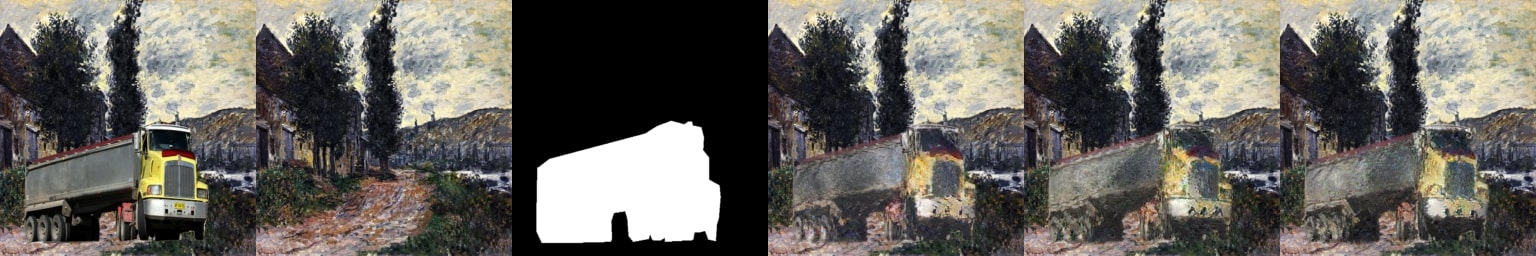}
    \end{subfigure}
    \begin{subfigure}{0.95\linewidth}
        \captionsetup{labelformat=empty,font=scriptsize}
        \begin{subfigure}{0.16\linewidth}
            \caption{composite}
        \end{subfigure}
        \begin{subfigure}{0.16\linewidth}
            \caption{background}
        \end{subfigure}
        \begin{subfigure}{0.16\linewidth}
            \caption{mask}
        \end{subfigure}
        \begin{subfigure}{0.16\linewidth}
            \caption{deep}
        \end{subfigure}
        \begin{subfigure}{0.16\linewidth}
            \caption{shallow}
        \end{subfigure}
        \begin{subfigure}{0.15\linewidth}
            \caption{Ours}
        \end{subfigure}
    \end{subfigure}
    \caption{The harmonization results obtained by adding residual features to different layers. }
    \label{fig:residual}
\end{figure}

By default, we add residual features to all four encoder layers, that is, $l=1,\ldots,4$.
In this section, we investigate the impact of adding learned residual features to only two shallow layers ($l=1,2$) or only two deep layers ($l=3,4$).
As shown in Figure \ref{fig:residual}, we observe that adding residual features only to partial layers may lose some detailed information (\emph{e.g.}, small letters on the stop sign in row 1, the front of the truck in row 3) or generate undesired artifacts (\emph{e.g.}, black spots on the chair in row 2), probably because some layers of feature maps are not well-harmonized. Instead, after adding residual features to all layers, our method can produce harmonized results with sharp details, smooth appearance, and reasonable colors, which demonstrates the effectiveness of modulating all layers of feature maps. 

\begin{figure*}[t]
  \centering
    \begin{subfigure}{\linewidth}
        \includegraphics[width=\linewidth]{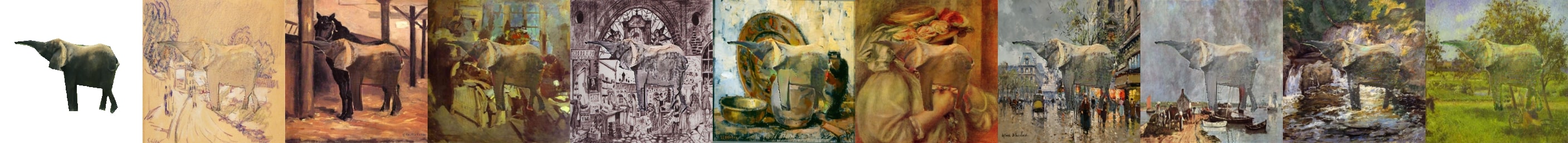}
    \end{subfigure}
    \begin{subfigure}{\linewidth}
        \includegraphics[width=\linewidth]{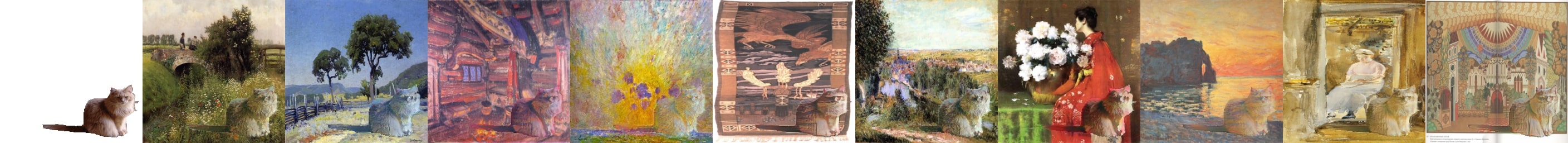}
    \end{subfigure}
    \begin{subfigure}{\linewidth}
        \includegraphics[width=\linewidth]{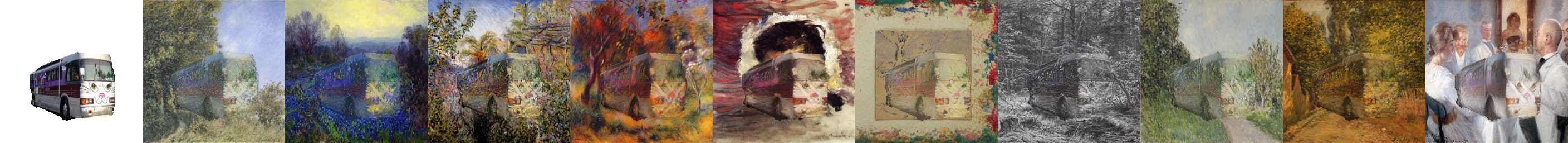}
    \end{subfigure}
    \begin{subfigure}{\linewidth}
        \includegraphics[width=\linewidth]{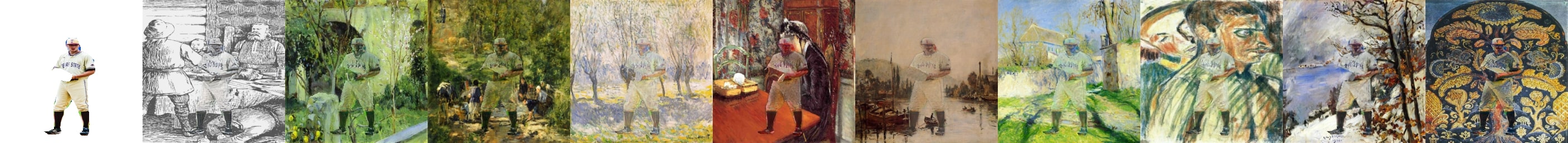}
    \end{subfigure}
    \begin{subfigure}{\linewidth}
        \includegraphics[width=\linewidth]{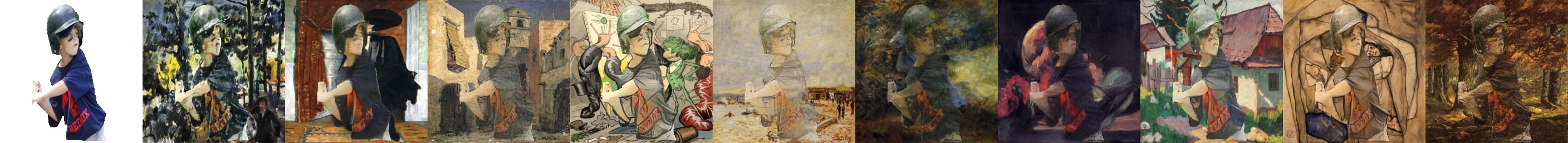}
    \end{subfigure}
    \begin{subfigure}{\linewidth}
        \captionsetup{labelformat=empty}
        \begin{subfigure}{0.087\linewidth}
            \caption{foreground}
        \end{subfigure}
        \begin{subfigure}{0.087\linewidth}
            \caption{1}
        \end{subfigure}
        \begin{subfigure}{0.087\linewidth}
            \caption{2}
        \end{subfigure}
        \begin{subfigure}{0.087\linewidth}
            \caption{3}
        \end{subfigure}
        \begin{subfigure}{0.087\linewidth}
            \caption{4}
        \end{subfigure}
        \begin{subfigure}{0.087\linewidth}
            \caption{5}
        \end{subfigure}
        \begin{subfigure}{0.087\linewidth}
            \caption{6}
        \end{subfigure}
        \begin{subfigure}{0.087\linewidth}
            \caption{7}
        \end{subfigure}
        \begin{subfigure}{0.087\linewidth}
            \caption{8}
        \end{subfigure}
        \begin{subfigure}{0.087\linewidth}
            \caption{9}
        \end{subfigure}
        \begin{subfigure}{0.087\linewidth}
            \caption{10}
        \end{subfigure}
    \end{subfigure}
  \caption{From left to right, we show the foreground object, the harmonized results of the same foreground pasted on different ten background pictures.}
  \label{fig:foreground}
\end{figure*}

\section{The Same Foreground on Different Backgrounds}
\label{sec:foreground}
We show the harmonized results when pasting the same foreground on different background images in Figure \ref{fig:foreground}.
We observe that with the preserved content structure, the foreground could be sufficiently stylized and harmonious with different backgrounds, which demonstrates the generalization ability of our method to cope with various combinations of foregrounds and backgrounds.

\section{Comparing Different Adversarial Losses}
\label{sec:adversarial}

We change our pixel-wise adversarial loss used for both encoder feature maps and output images to the vanilla adversarial loss in \cite{peng2019element} and the domain verification adversarial loss in \cite{cong2020dovenet}, while keeping the other network components unchanged. 
The adversarial losses in \cite{peng2019element} and \cite{cong2020dovenet} represent image-wise adversarial loss and region-wise adversarial loss respectively. Therefore, we actually compare three types of adversarial losses: image-wise, region-wise, and pixel-wise adversarial losses. 

We show the visual comparison below, which demonstrates that pixel-wise adversarial loss performs far better than other types of adversarial losses.
Additionally, we invite 50 users to select from three methods for 100 composite images, which shows that 87\% users choose our method.

\begin{figure}[t]
    \centering
    \begin{subfigure}{0.95\linewidth}
        \includegraphics[width=\linewidth]{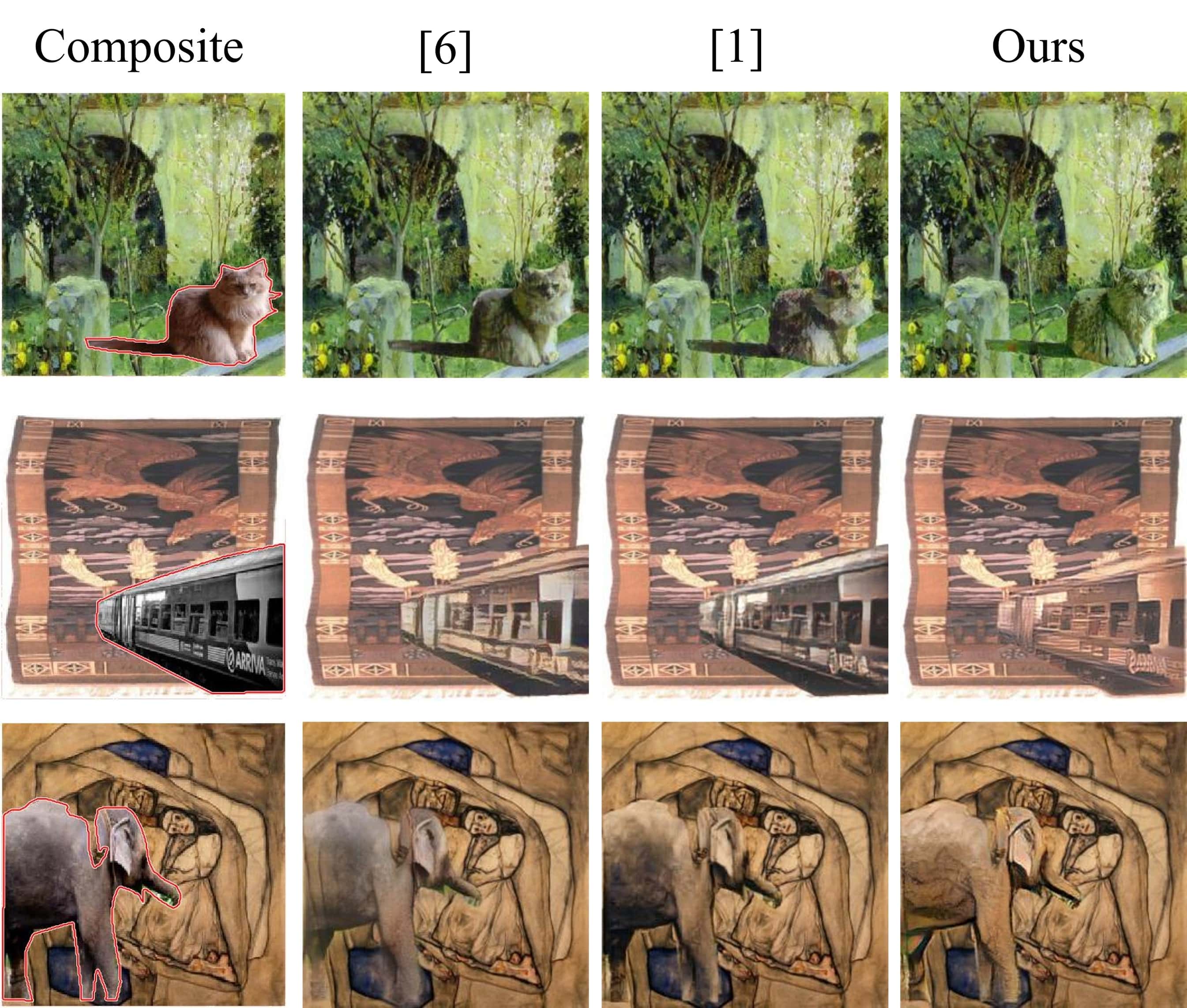}
    \end{subfigure}
    \caption{From left to right, we show the composite image, the harmonized results of using adversarial loss of \cite{peng2019element}, \cite{cong2020dovenet} and our method.}
    \label{fig:failure}
\end{figure}

\section{Multiple Foregrounds on One Background}
\label{sec:multifore}

\begin{figure}[t]
    \centering
    \begin{subfigure}{0.95\linewidth}
        \includegraphics[width=\linewidth]{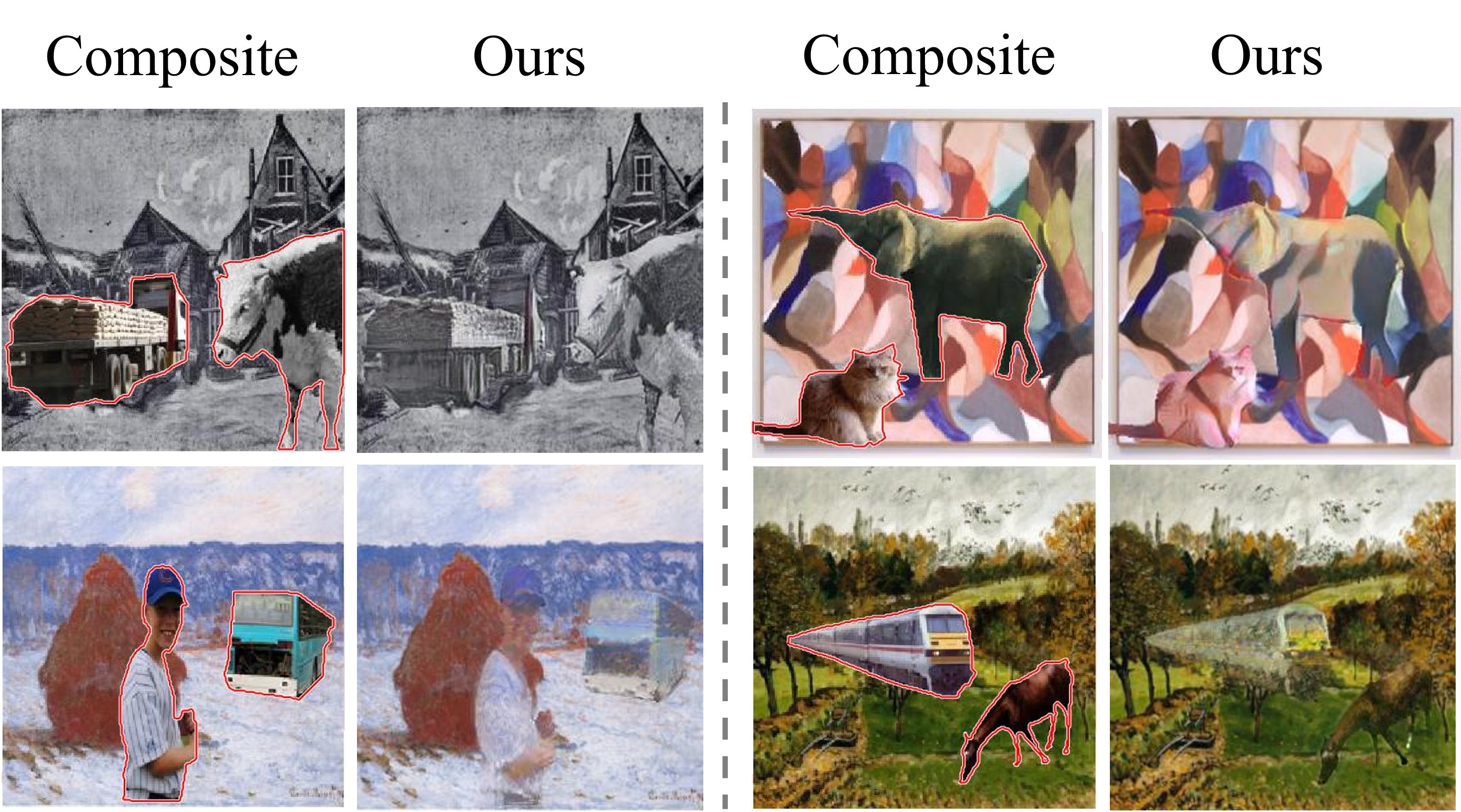}
    \end{subfigure}
    \caption{Example of multiple foregrounds on one background.}
    \label{fig:multi_foreground}
\end{figure}

Our method can be directly applied to the test images with multiple composite foregrounds. 
We can just feed the composite image and mask with multiple foregrounds, passing through the network once. We show some results in Figure~\ref{fig:multi_foreground}, which shows that our method can harmonize multiple foregrounds simultaneously. 

{\small
\bibliographystyle{ieee_fullname}
\bibliography{supp.bbl}
}